# An intelligent approach towards automatic shape modeling and object extraction from satellite images using cellular automata based algorithm

P. V. Arun[1] S.K. Katiyar

MANIT-Bhopal

India

**Abstract**

Automatic feature extraction domain has witnessed the application of many intelligent methodologies over past decade; however detection accuracy of these approaches were limited as object geometry and contextual knowledge were not given enough consideration. In this paper, we propose a frame work for accurate detection of features along with automatic interpolation, and interpretation by modeling feature shape as well as contextual knowledge using advanced techniques such as SVRF, Cellular Neural Network, Core set, and MACA. Developed methodology has been compared with contemporary methods using different statistical measures. Investigations over various satellite images revealed that considerable success was achieved with the CNN approach. CNN has been effective in modeling different complex features effectively and complexity of the approach has been considerably reduced using corset optimization. The system has dynamically used spectral and spatial information for representing contextual knowledge using CNN-prolog approach. System has been also proved to be effective in providing intelligent interpolation and interpretation of random features.

[1] Email : arunpv2601@gmail.com

**Keywords:** Cellular Automata; Remote Sensing; Object Extraction

## 1. Introduction

Detection and identification of objects from satellite images is a prerequisite of many applications; however, accurate and robust detection of objects remain a problem because of various factors, such as geometrical complexity noise, vague boundaries, mixed pixel problems, and fine characteristics of detailed structures (Daniel, 2006; Jacek, 2004). Different existing algorithms are specific to features to be extracted and adopt computationally complex methods for object extraction (Yuan et al., 2009; Sunil, 2004). The efficiency of these methods are situation- and image-specific due to involvement of various parameters like spatial and spectral resolution, sensor characteristics, etc. (Trinder et al., 2003; Mnih et al., 2010).

There has been a long history of the usage of intelligent methodologies in the context of object extraction. Soft computing techniques, such as neural networks, genetic algorithms, and fuzzy logic followed by probabilistic concepts such as random field variations, have been extensively applied in this context (Lari et al., 2011; Wang et al., 2009; Chi et al, 2004). The literature has also revealed many object enhancement filters as well as intensity-based approaches (Jensen et al., 2003; Haralick et al., 1983). N-dimensional classifiers as well as random field concepts and different transformation techniques have been applied for accurate detection (Hosseini et al., 2009; Sanjiv et al., 2003; Chang et al., 2006). Contextual information is a key factor for real-time detection to avoid ambiguity; knowledge-based classification approaches such as predicate calculus have been recently used in this context (Porway et al., 2008; Harvey, 2004).

Different image interpretation features namely tone, texture, pattern and colour were employed in most of above mentioned approaches. Modelling of shape is being less exploited in the context of feature detection and is a major factor in distinguishing the entities (Lindi, 2004; Harvey, 2004). Our studies have found that inverse mapping of Cellular Automata using genetic algorithm can be efficiently used for modeling shape of features. Literature review of works in this field also revealed that cellular automata have been proven to be an efficient method for image enhancement and classification (Orovas et al, 1998; Mitchell et al, 1993). This approach seems to improve feature detection accuracies to a great extent as revealed from our investigation.

Predicate calculus rule sets used for context representation fail to represent spatial relations effectively as it lack an image compatible form (Mitchell et al, 1993). Cellular automata rules can be used as an alternative as it can represent image rules more effectively (Orovas et al, 1998). Rule sets can be set by using a neural net, thus generating a cellular neural network for the purpose. SVM approaches are supervised and hence may not help for clustering (Huang et al, 2002) however an adaptive kernel strategy introduced by Srivastava (2004) may help in exploiting n-dimensional classifier concept for achieving an unsupervised strategy. Main obstruct in the modelling of features using CNN approach is increased computational complexity, which can be effectively tackled using an aproximation by CORESET.

In this paper we present an object extraction methodology in which cellular neural network is used for modelling feature shape, and adaptive kernel strategy along with corset optimisation are adopted to make the technique unsupervised. Salient features of this work are cellular automata approach based on adaptive kernel strategy, intelligent interpolation and automatic interpretation. Accuracy of developed methodologies are compared with contemporary approaches using satellite images of Bhopal and Chandrapur cities in India

## 2. Theoretical Back Ground

### *2.1 CNN*

Cellular Neural Network (CNN) (Orovas et al, 1998; Mitchel et al, 1993) is an analog parallel computing paradigm defined in space and characterized by locality of connections between processing elements (cells or neurons). An uncoupled CNN can be described as a simple non linear equation as it converges towards a steady state output over a longer time interval. Cell dynamics of this continuous dynamic system may be denoted using ODEs as $\boldsymbol{X1}\ (\boldsymbol{t}) = -\boldsymbol{X1} + \sum_{k \in N} \mathbf{a_k}\, \mathbf{Y_k} + \sum_{k \in N} \mathbf{b_k}\, \mathbf{u_k} + \mathbf{Z}$ and output is calculated as $\mathbf{Y_k}\ (\boldsymbol{t}) = \frac{1}{2}(|\boldsymbol{X_k} + \mathbf{1}| - |\boldsymbol{X_k} + \mathbf{1}| - |\boldsymbol{X_k} - \mathbf{1}|$. Thus the non standard cell equation can be stated as $\mathrm{X_k}\ (\mathrm{t}) = -\mathrm{X1} + \mathrm{f}(\mathrm{G}, \mathrm{Y_k}, \mathrm{U_K})$where vector G is the gene which determines random nature. Like cellular automata, each cell communicate with other directly only through its nearest neighbours and indirectly due to propagation effects of continuous-time dynamics of cellular neural networks.

### *2.2 MACA*

Multiple Attractor Cellular Automata is a special type of CA in which different local rules are applied to different cells and MACA will get converged to certain attractor states on execution (Weunsh et al, 2003; Sikdar et al, 2000). In an n cell MACA with $2^m$ attractors, there exist m-bit positions at which attractors give pseudo-exhaustive 2m patterns. In order to identify the class of a pattern, MACA is initialized with the pattern and operated for maximum (depth) number of cycles till it converges to an attractor and then PEF bits are extracted from attractor to identify class of that pattern.

***2.3 Mixed density kernel***

Mercer Kernel functions can be viewed as a measure of similarity between two data points that are embedded in a high, possibly infinite dimensional feature space. A *mixture density kernel* (Srivastava, 2004) is a Gram matrix that measures the number of times an ensemble of mixture density estimates agree that two points arise from same mode of probability density function and is given as

$$K(X_i, X_j) = \frac{1}{Z(X_i, X_j)} \sum_{m=1}^{M} \sum_{cm=1}^{Cm} P_m\left({C_m}/{X_i}\right) P_m\left({C_m}/{X_j}\right)$$

Where 'M' be the number of clusters and P $(C_m/X_i)$ be the probability that data point '$X_i$' belongs to $C_m$.

### 2.4 SVRF

SVRF (Schnitzspan et al, 2008) is a Discrete Random Field (DRF) based extension for SVM, constituting of observation-matching potential and local-consistency potential functions. The observation-matching function captures relationships between observations and class labels, while the local-consistency function models relationships between labels of neighboring data points and observations at data points. SVRF is described as,

$$P(Y \mid X) = \frac{1}{Z} \exp\left\{ \sum_{i \in S} \log(O(y_i, \Gamma_i(X))) + \sum_{i \in S} \sum_{j \in N_i} V(y_i, y_j, X) \right\}$$

Where, $\Gamma_i(X)$ is a function that computes features from observations *X* for location *i*, *O(yi, i(X))* is an SVM-based Observation-Matching potential and *V (yi, yj ,X)* is a (modified) DRF pair wise potential.

### 2.5 Coreset

Coreset (Agarwal et al, 2001; Badoiu et al, 2002) is small subset of a point set, which is used to compute a solution that approximates solution of entire point set. Let μ be a measure function (e.g., width of a point set) from subsets of $R^d$ to non-negative reals $R^+U\{0\}$ that is monotone, i.e., for P1 $C$ P2, μ(P1) ≤ μ(P2). Given a parameter $\varepsilon > 0$, we call a subset Q $C$ P as an $\varepsilon$ -Coreset of P (with respect to μ) if (1 $-\varepsilon$) μ (P) ≤ μ (Q).

## 3. Methodology

### *3.1 Clustering*

Initial clustering is accomplished by using Support Vector Random Field approach that uses mixture density kernel. Adopted mixture density kernel includes contextual information by incorporating both spectral and spatial information using CA rules. Composite kernel concept is used to incorporate spectral and spatial information, i.e. given $X=\{x_1,x_2,..x_m\}^T$ be spectral characteristics of an M-band multispectral imagery and $Y=\{y_1,y_2,..y_n\}^M$ be spatial characteristics, then the possible spectral and spatial kernels can be denoted as $K_x(P,P_i) = \langle\Phi(P), \Phi(P_i)\rangle$ , $K_y(P,P_i) = \langle\Psi(P), \Psi(P_i)\rangle$ respectively. Preferably a weighted combination of kernels are adopted as discussed in [5] such that $K(P,P_i) = \mu K_x(P,P_i) + (1-\mu)K_y(P,P_i)$ and value of tuning parameter is adjusted accordingly.

Kernel function counts the number of times ‘$M$ mixtures’ agree to place two points in same cluster mode. Further interpretation of mixed density kernel function using Bayes rule shows that Mixture Density Kernel measure the ratio of probability that two points arise from same mode, compared with unconditional joint distribution.

$$K(X_i, X_j) = \frac{1}{Z(X_i, X_j)} \sum_{m=1}^{M} \sum_{cm=1}^{Cm} \frac{P_m{}^2\left({X_i, X_j}/{C_m}\right) P_m{}^2(C_m)}{P_m(X_i)}$$

If we simplify this equation further by assuming that class distributions are uniform, kernel tells us the amount of information gained by knowing that two points are drawn from same mode. Thus parameters of mixture density kernels are adjusted automatically based on ensembles, and hence can be exploited to incoporate contextual information as well as adaptive kernel strategy for SVRF.

### *3.2 Object extraction*

Once the edges are detected, CA based region growing strategy is adopted to extract objects. Each pixel is assigned a state, namely 'B' for boundary pixel, 'NB' for non boundary pixel and 'NR' for Snon region pixels. Initially boundary pixel states are assigned as 'B' and non boundary pixel sates as 'UB'. The 'NB' pixel's state is changed to 'NR' iteratively if it is near to boundary pixel. The whole procedure is repeated until no further state change is experienced, hence detecting different objects in the image.

### *3.3 Coreset optimisation*

Objects in satellite images may span over a long distance and hence a reduction in pixel number (n) is needed to accomplish object modeling in acceptable complexity range. Approximation of features using a corset based approach will help to reduce the number of pixels considerably without losing original shape. Let *S* be a corset for **a** corset scheme *A,* F be the class of queries, P be finite set of points, evolve be a function and $\varepsilon$ be a parameter such that $\varepsilon \leq 0$ & $\varepsilon \leq 0.49$. Let the output $A(P) = S$ *is* such that for every *F*, $(1 - \varepsilon) \cdot evolve\ (P, F) \leq evolve\ (A(P),F) \leq (1 + \varepsilon)\ evolve\ (P,F)$. *Then S is said to be a corset and can be adopted for approximation of features without affecting results of inverse modelling.*

For a given Feature *F,* a line corset ($k$, ε) is constructed so that it resembles *F* with much lesser number of pixels, $K<<F$. Given any *n* points in $R^d$, a ($k$, ε) line coreset of size polygon $\log(n)^x$ can be constructed in $nd.k^{O(1)} = O(n)$ time, given $x = 2^{O(k)} (\frac{1}{\varepsilon})^{d+2k} \log^{4k}$ (n). A ($1+\varepsilon$) approximation algorithm for the *k*-line median problem ($O(n)$ time) is also available how ever due to accuracy concern trivial approach was selected.

### *3.4 Object interpretation*

CNN along with GA can be effectively used to find rules that iterates from a given initial state to a desired final state. This inverse mapping or evolution is exploited to model feature shapes, and CNN rules used to evolve a particular feature is used to distinguish it. Rules corresponding to various features are thus deducted and are stored in a prolog DB. In addition to feature interpretation, these rules are also used to guide mutation and crossover of GA to increase efficiency. Entropies and dimensions give a generalized measure for configurations generated by cellular automaton evolution. The (set) dimension or limiting (topological) entropy for a set of cellular automaton configurations is defined as

$d^{(x)} = \lim_{X \to \infty} \frac{1}{X} \log_k N(x)$,where $N(X)$ gives the number of distinct sequences of X site values that appear. Thus inverse evolution can be attained in lesser than $n \log(n)$time, provided that the features will converge to lower class CA configurations. Core set based optimization is used for feature approximation so that features can be effectively mapped to lower class CA configurations. MACA is automatically initialised with most likely pattern, to identify the class of the pattern in less than $\log(n)$ time.

### *3.5 Intelligent interpolation*

Interpolations of features such as road, rivers are accomplished by using Cellular Automata rules integrated with predicate rules. For e.g., given a feature like road, training experience or stored rule is used to set the threshold value for particular evolution rule to be adopted. Once a pattern for an object is found and stored in prolog DB, next time onwards core set based optimisation will be utilised to estimate likeness of an object to that pattern. Then MACA is automatically initialised with the most appropriate pattern to identify class of interpolated pattern in less than $\log(n)$ time.

## 4. Results

Investigations of feature extraction process over various satellite images revealed that considerable success was achieved with CNN approach. This approach accurately detects various features better than existing methodologies and accuracy was revealed by extraction of air strip, roads and rivers. Efficiency of traditional classifying approaches with reference to CNN approach has been evaluated using various statistical measures and results are as summarised in (Table 2). Ground truthing have been done by virtue of Google earth and Differential Global Positioning System (DGPS) survey over the study area using Trimble R3 DGPS equipment. Increased values of Kappa statistics and over all accuracy indicates better method.

Table 2. Comparative analysis

| S.No | Sensor | Methodology | Kappa statistics | Overall Accuracy (%) |
|---|---|---|---|---|
| 1 | LISS 3 | Mahalanobis | 0.91 | 92.13 |
| 2 | LISS 3 | Minimum Distance | 0.92 | 93.58 |
| 3 | LISS 3 | Maximum Likelihood | 0.93 | 94.83 |
| 4 | LISS 3 | Parrellelepipid | 0.94 | 95.81 |
| 5 | LISS 3 | Feature Space | 0.95 | 95.15 |
| 6 | LISS 3 | CNN Approach | 0.98 | 98.82 |
| 8 | LISS 4 | Mahalanobis | 0.90 | 91.40 |
| 9 | LISS 4 | Minimum Distance | 0.91 | 93.00 |
| 10 | LISS 4 | Maximum Likelihood | 0.94 | 94.80 |
| 11 | LISS 4 | Parrellelepipid | 0.93 | 94.62 |
| 12 | LISS 4 | Feature Space | 0.94 | 95.31 |
| 13 | LISS 4 | CNN | 0.98 | 97.82 |

Performances of these methodologies have also been evaluated by comparing areal extents of various features. Features namely lakes and parks were selected for comparative analysis, since these features are having well defined and fixed geometry which is distinguishable using most of recent methods. Original surface areas of various extracted features are calculated by manual digitization using ERDAS and comparative analysis of results is presented in (Table 3). Comparative analyses of areal extents also indicate that CNN approach yields better results compared to other methods.

Table 3. Comparative analysis of areal accuracy

| S.No | Sensor | Feature | Reference Area($km^2$) | Methodology | Areal Extent($km^2$) |
|---|---|---|---|---|---|
| 1 | LISS3 | Lake | 32.5 | Mahalanobis | 25.42 |
| | | | | Minimum Distance | 24.31 |
| | | | | Maximum Likelihood | 27.37 |
| | | | | Parallelepiped | 28.58 |
| | | | | Feature Space | 26.82 |
| | | | | CNN Approach | 28.01 |
| | | | | Mahalanobis | 0.82 |
| | | | | Minimum Distance | 0.89 |
| 2 | LISS3 | Parks | 2.13 | Maximum Likelihood | 1.45 |
| | | | | Parallelepiped | 1.37 |
| | | | | Feature Space | 1.56 |
| | | | | CNN Approach | 2.07 |
| 3 | LISS4 | Lake | 32.81 | Mahalanobis | 24.31 |
| | | | | Minimum Distance | 23.40 |
| | | | | Maximum Likelihood | 25.12 |
| | | | | Parallelepiped | 26.24 |
| | | | | Feature Space | 27.17 |
| | | | | CNN | 28.01 |
| 4 | LISS4 | Parks | 2.37 | Mahalanobis | 0.51 |
| | | | | Minimum Distance | 0.72 |
| | | | | Maximum Likelihood | 1.53 |
| | | | | Parallelepiped | 1.14 |
| | | | | Feature Space | 1.46 |
| | | | | CNN | 1.62 |

The visual results of few features extracted by system are given below, which also reveals the accuracy of CNN approach.

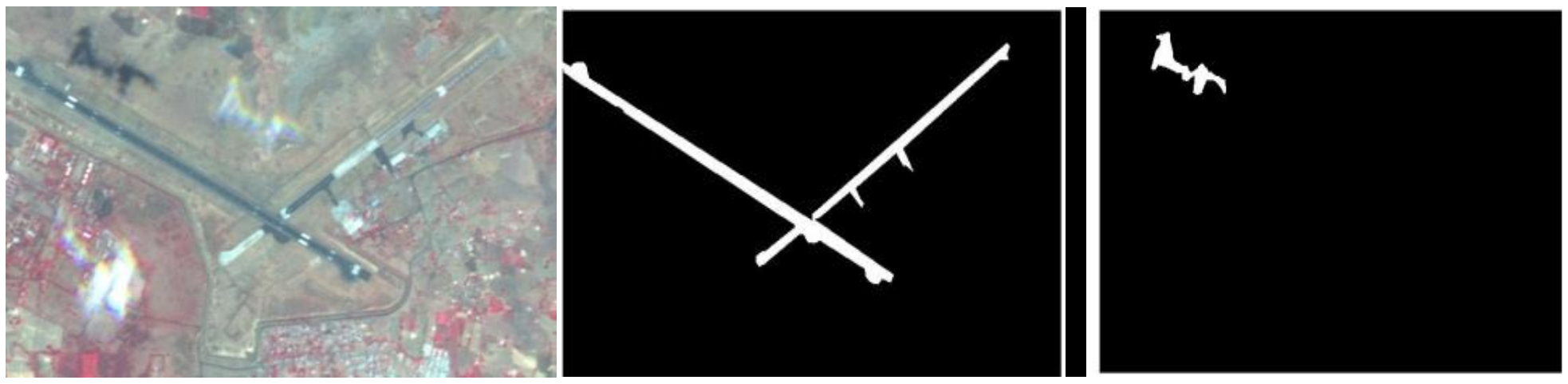

**a**.)Original image b) Road network c) Cloud

Figure 2. Features extracted from LISS 3 sensor image

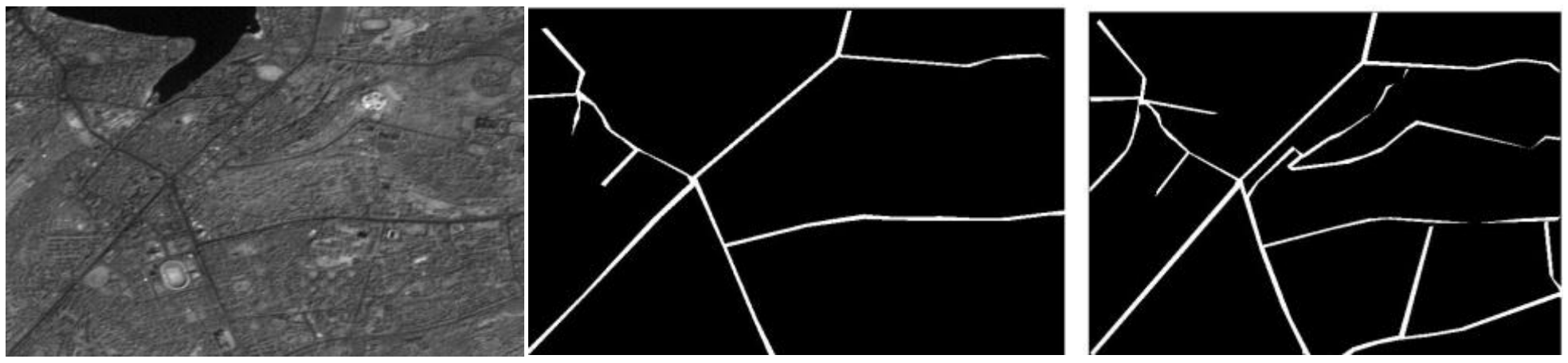

a.)Original image b) Road network c) Road Network

Figure 3. Road network extracted from PAN sensor image

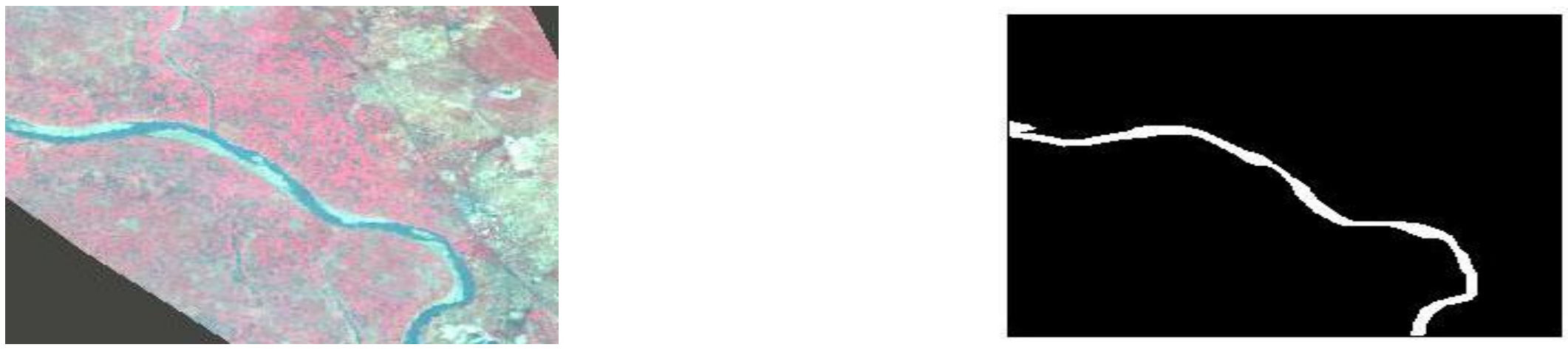

a.)Original image b) River

Figure 4. River extracted from LANDSAT sensor image

## 5. Conclusion

We discussed a CNN based approach that could effectively model feature shapes and context sensitivity. Investigations have revealed that the method outperforms contemporary approaches in terms of accurate detection. Paper provides a frame work for CA based feature shape modeling. Complexity of the approach has been considerably reduced using corset based approximation. Proposed system has proved to be intelligent with reference to accurate interpolation and interpretation. Disambiguations of features, enhanced detection, self learning, minimal human interpretation, reliability are features of the system.